\documentclass[iicol,sn-mathphys,Numbered]{sn-jnl}

\usepackage{graphicx}%
\usepackage{multirow}%
\usepackage{amsmath,amssymb,amsfonts}%
\usepackage{amsthm}%
\usepackage{mathrsfs}%
\usepackage[title]{appendix}%
\usepackage{xcolor}%
\usepackage{textcomp}%
\usepackage{manyfoot}%
\usepackage{booktabs}%
\usepackage{algpseudocode}%
\usepackage{listings}%
\usepackage{booktabs}
\usepackage{multicol}
\usepackage{multirow}
\usepackage{subcaption}
\usepackage[ruled,linesnumbered,vlined]{algorithm2e}
\usepackage{color, colortbl}
\usepackage{xcolor}

\theoremstyle{thmstyleone}%
\theoremstyle{thmstyletwo}%

\theoremstyle{thmstylethree}%

\DeclareMathOperator*{\argmax}{arg\,max}

\begin{document}

\title[Article Title]{Color Prompting for Data-Free Continual Unsupervised Domain Adaptive Person Re-Identification}


\author[1]{\fnm{Jianyang} \sur{Gu}}\email{gu\_jianyang@zju.edu.cn}

\author[2]{\fnm{Hao} \sur{Luo}}\email{haoluocsc@zju.edu.cn}

\author[3]{\fnm{Kai} \sur{Wang}}\email{e0823044@u.nus.edu}

\author*[1]{\fnm{Wei} \sur{Jiang}}\email{jiangwei\_zju@zju.edu.cn}

\author[3]{\fnm{Yang} \sur{You}}\email{youy@comp.nus.edu.sg}

\author*[4]{\fnm{Jian} \sur{Zhao}}\email{zhaojian90@u.nus.edu}


\affil*[1]{\orgdiv{College of Control Science and Engineering}, \orgname{Zhejiang University}, \orgaddress{\city{Hangzhou}, \country{China}}}

\affil[2]{\orgname{Alibaba Group}, \orgaddress{\city{Hangzhou}, \country{China}}}

\affil[3]{\orgdiv{School of Computing}, \orgname{National University of Singapore}, \orgaddress{\country{Singapore}}}

\affil[4]{\orgname{Institute of North Electronic Equipment}, \orgaddress{\city{Beijing}, \country{China}}}


\abstract{Unsupervised domain adaptive person re-identification (Re-ID) methods alleviate the burden of data annotation through generating pseudo supervision messages. However, real-world Re-ID systems, with continuously accumulating data streams, simultaneously demand more robust adaptation and anti-forgetting capabilities. Methods based on image rehearsal addresses the forgetting issue with limited extra storage but carry the risk of privacy leakage. In this work, we propose a Color Prompting (CoP) method for data-free continual unsupervised domain adaptive person Re-ID. Specifically, we employ a light-weighted prompter network to fit the color distribution of the current task together with Re-ID training. Then for the incoming new tasks, the learned color distribution serves as color style transfer guidance to transfer the images into past styles. CoP achieves accurate color style recovery for past tasks with adequate data diversity, leading to superior anti-forgetting effects compared with image rehearsal methods. Moreover, CoP demonstrates strong generalization performance for fast adaptation into new domains, given only a small amount of unlabeled images. Extensive experiments demonstrate that after the continual training pipeline the proposed CoP achieves 6.7\% and 8.1\% average rank-1 improvements over the replay method on seen and unseen domains, respectively.  }

\keywords{Person Re-Identification, Color Style Transfer, Unsupervised Learning, Continual Learning}



\maketitle

\section{Introduction}
Person Re-Identification (Re-ID) aims to retrieve images of specific pedestrian identities across cameras from a large gallery given a query image~\cite{ye2021deep,yi2014deep,zheng2016person,li2014deepreid,luo2019strong,wu2018and,yan2021beyond}. 
Recent years have witnessed significant progresses of person Re-ID owing to the rapid development of deep neural networks~\cite{luo2019alignedreid++,hou2019vrstc,wu2017rgb,miao2019pose}. 
However, the success of previous supervised Re-ID methods heavily relies on extensive data annotation efforts. 
Consequently, there has been a surge in the development of unsupervised methods, aiming to alleviate the heavy cost of the annotation process~\cite{ge2020mutual,ge2020self,zou2020joint,song2020unsupervised}. 

Despite achieving considerable accuracy given a specific domain, there still exists non-negligible gap between current unsupervised person Re-ID systems and the practical application. 
The disparity is primarily resulted from the explosive influx of the data collection process carried by widely installed surveillance systems~\cite{pu2021lifelong,wu2021generalising}. 
As a result, a trained model can easily become inadequate as the data domain shifts over time. 
Nevertheless, fine-tuning the model with all captured data is impractical and unaffordable. 
Consequently, there occurs an urgent need for continual learning methods that can introduce anti-forgetting capabilities for previous data and enhance the performance on the newly-collected data~\cite{wu2021generalising,ge2022lifelong,huang2022lifelong}. 

There have been several works proposed to address the problem of continual Re-ID, which can be broadly categorized into regularization-based and replay-based methods. 
Regularization-based methods constrain the parameter updates of the model when training on new tasks~\cite{pu2021lifelong,zhao2021continual,sun2022patch}. 
While the forgetting on the past data is reduced, the direct constraints can lead to sub-optimal adaptation performance on new domains. 
On the other hand, replay-based methods acquire the access to past knowledge through maintaining an auxiliary image memory of data from previous tasks~\cite{chen2022unsupervised,huang2022lifelong,wu2021generalising}. 
However, the accessibility to the past data can raise concerns on privacy leakage issues. 
Prior researches have explored both the supervised and unsupervised scenarios in the context of continual Re-ID. 
In this work, we mainly target on the more practical and challenging continual unsupervised domain adaptive (UDA) person Re-ID task. 

We propose to design a light-weighted data-free approach to tackle the continual UDA person Re-ID problem. 
We first analyze the color distribution gap existing between different cameras and domains. 
Accordingly, we propose a Color Prompting (CoP) method to mitigate the domain bias from the perspective of color styles. 
Specifically, along with the source domain supervised training, we employ a compact prompter network to capture the color style distribution. 
When applied to the target domain, the prompter predicts the color distribution based on the contents of each image. 
Then a color style transfer process is conducted to generate images with the color style of the source domain. 
Through this process, we effectively re-obtain the knowledge from previous tasks without explicitly storing the data. 

Our experimental findings validate that CoP achieves superior anti-forgetting effects without storing images of previous tasks, compared with replay-based methods. 
Furthermore, the increased data diversity introduced by the color style transfer also leads to improvements in the adaptation performance on the target domain and the generalization capabilities. 
Overall CoP achieves 6.7\% and 8.1\% average rank-1 improvements over the replay method on seen and unseen domains, respectively. 
Besides, given a small amount of unlabeled images from a specific domain, CoP demonstrates the ability to fast adapt the model to acquire considerable performance on the target domain. 
The characteristic makes CoP even more practical for real-world application of unsupervised Re-ID systems. 

The contribution is summarized as follows,
\begin{itemize}
    \item We analyze the color distribution gap between different cameras and domains.
    \item We propose a Color Prompting (CoP) framework to create past knowledge rehearsal without explicitly storing data of previous tasks.
    \item We conduct extensive experiments to verify that CoP is an effective, efficient and privacy-friendly method for continual UDA person Re-ID, where 6.7\% and 8.1\% average rank-1 improvements are achieved over replay method. \footnote{The source code for this work is publicly available in \href{https://github.com/vimar-gu/ColorPromptReID}{https://github.com/vimar-gu/ColorPromptReID}.}
\end{itemize}

\section{Related Works}
\subsection{Person Re-Identification}
Person Re-Identification (Re-ID) is a fine-grained branch of metric learning tasks~\cite{ye2021deep,yi2014deep,zheng2016person,li2014deepreid,luo2019strong,wu2018and,yan2021beyond}. 
Given a query image, it aims to retrieve similar images from a large gallery across different cameras. 
Along with the development of city surveillance systems, Re-ID has become an essential tool for building up intelligent security and smart retail systems. 
There have been abundant works focusing on various person Re-ID settings, such as standard image-based scenario~\cite{luo2019strong,luo2019alignedreid++}, video-based scenario~\cite{hou2019vrstc,mclaughlin2016recurrent,li2019global}, cross modality scenario~\cite{wu2017rgb,ye2018hierarchical,feng2023shape}, occlusion scenario~\cite{miao2019pose,zheng2015partial,he2019foreground} and clothe-changing scenario~\cite{yang2019person,gu2022clothes,yu2020cocas}. 
There are also some literature focusing on designing better architectures for person Re-ID~\cite{zhou2019omni,quan2019auto,gu2023msinet}. 
Sufficient data annotation helps boost the recognition accuracy of supervised person Re-ID methods, but also demands large amounts of labor and time cost. 
Unsupervised methods aim at providing considerable model performance without further annotation on the incoming new data domains. 

\subsection{Unsupervised Person Re-Identification}
Unsupervised methods have been widely applied to the person Re-ID task to mitigate the dependence on the data annotation. 
Based on the dependency on the source domain knowledge, previous works can be divided into purely unsupervised methods~\cite{dai2022cluster,fan2018unsupervised,lin2019bottom,yu2019unsupervised,wang2020unsupervised} and unsupervised domain adaptation methods~\cite{wei2018person,song2020unsupervised,ge2020self,gu2022multi}. 
In this work, we mainly focus on the more practical domain adaptation setting. 
Previous works introduce adversarial training to transfer the image style of annotated source data to specific domains, where the supervision is accurate. 
Wei \textit{et al.} propose to transfer the image style of source domain to the target domain~\cite{wei2018person}. 
Deng \textit{et al.} explicitly constrain the identity consistency during the generation process~\cite{deng2018image}. 
Zou \textit{et al.} disentangle the features into identity-related and unrelated parts for generating images with higher quality~\cite{zou2020joint}. 
Chen \textit{et al.} additionally modify the background of images to better simulate the target domain information~\cite{chen2019instance}. 
Besides, clustering algorithms are employed to generate pseudo labels for supervision~\cite{song2020unsupervised}. 
Although at the beginning the correctness of the supervision is poor, as the training processes, the model gradually generates more accurate labels. 
Ge \textit{et al.} introduce a mutual mean teaching scheme to improve the pseudo label quality~\cite{ge2020mutual}. 
Ge \textit{et al.} further propose a self-paced sample selection mechanism and a contrastive objective for proper supervision~\cite{ge2020self}. 
Some other works also focus on improving the reliability of the pseudo labels~\cite{zhao2020unsupervised,li2020joint,feng2021complementary}. 

However, previous methods primarily focus on the adaptation performance on target datasets, while lacking detailed consideration on more realistic real-world application scenarios. 
When the model is adapted to newly collected data, it is inevitable to forget previously accumulated knowledge. 
As fine-tuning the model on all the data can be prohibitively expensive, it is essential for a Re-ID system to minimize forgetting with the least dependence on accessing past data. 

\begin{figure*}[t]
\begin{subfigure}[t]{0.58\textwidth}
    \centering
    \includegraphics[width=\textwidth]{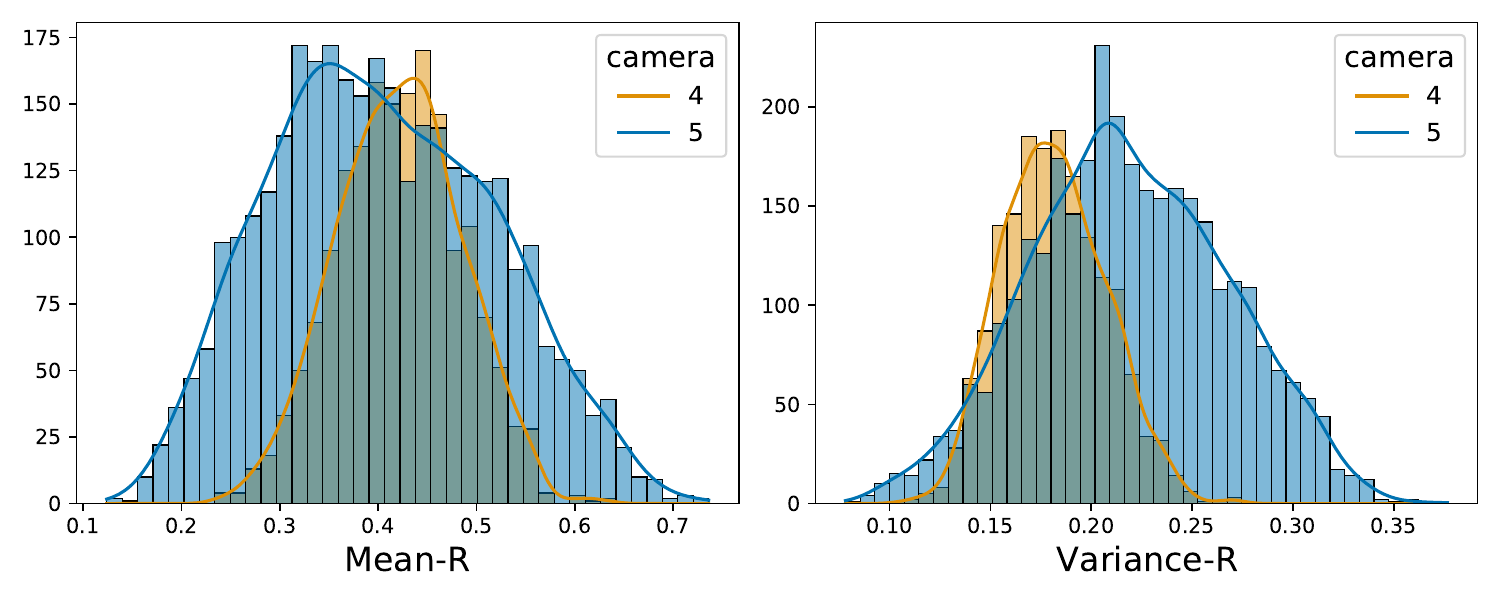}
    \caption{The red color distribution comparison between images captured by camera 4 and camera 5 in the dataset Market-1501. }
    \label{fig:hist-intra}
\end{subfigure}
\hfill
\begin{subfigure}[t]{0.38\textwidth}
    \centering
    \includegraphics[width=0.98\textwidth]{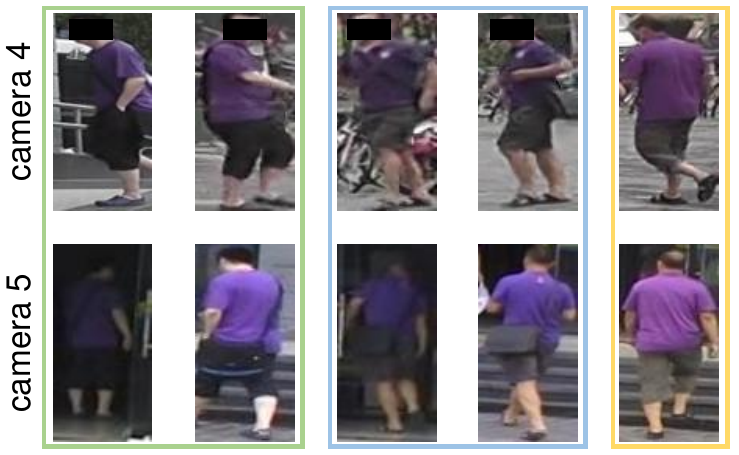}
    \caption{Example images captured by camera 4 and 5 in the dataset Market-1501.}
    \label{fig:example-intra}
\end{subfigure}

\begin{subfigure}[t]{0.59\textwidth}
    \centering
    \includegraphics[width=\textwidth]{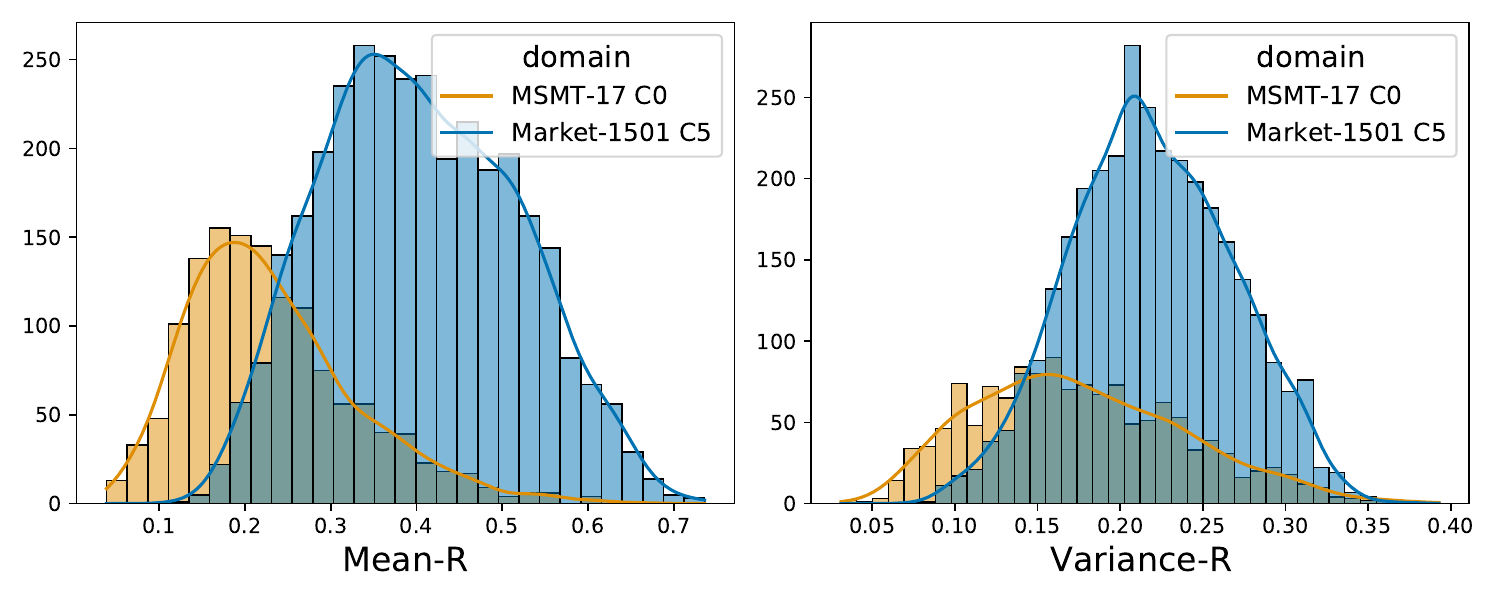}
    \caption{The red color distribution comparison between images in the dataset MSMT-17 camera 0 and Market-1501 camera 4. }
    \label{fig:hist-inter}
\end{subfigure}
\hfill
\begin{subfigure}[t]{0.38\textwidth}
    \centering
    \includegraphics[width=0.98\textwidth]{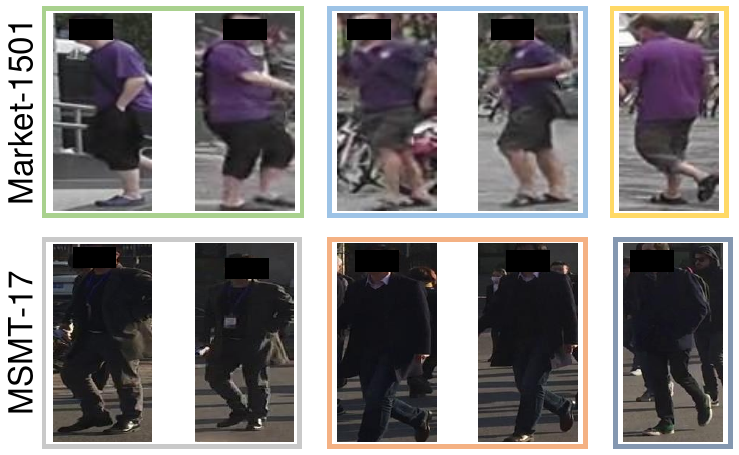}
    \caption{Example images in the dataset MSMT-17 camera 0 and Market-1501 camera 4.}
    \label{fig:example-inter}
\end{subfigure}

\caption{The color distribution gap between (a) cameras inside a system; (c) cameras of different systems and their corresponding example images (b), (d). There are distinguishable gap between different distributions. Different box colors in (b) and (d) refer to different identities. }
\end{figure*}

\subsection{Continual Learning}
Continual learning aims to incrementally accumulate knowledge from a series of non-overlapped tasks. 
As the access to all the data is limited, the model can be easily inadequate for previous tasks, which is denoted as \textit{catastrophic forgetting}. 
Recent continual learning methods can be roughly divided into dynamic model architecture~\cite{aljundi2017expert,fernando2017pathnet,kang2022forget,rosenbaum2018routing}, regularization~\cite{kirkpatrick2017overcoming,lee2017overcoming,li2017learning,zenke2017continual}, pseudo sample generation~\cite{shin2017continual,smith2021always,xiang2019incremental,zhu2021prototype}, and sample rehearsal~\cite{lopez2017gradient,rebuffi2017icarl,tiwari2022gcr,yoon2022online} methods. 

There are also several continual learning works focusing on the person Re-ID scenario.
Previous works mainly adopt regularization and replay techniques to tackle the catastrophic forgetting problem. 
Wu \textit{et al.} constrain the consistency of the feature representation and the predicted logits between old and new models~\cite{wu2021generalising}. 
Zhao \textit{et al.} design a neighbourhood selection and consistency relaxation strategy to improve the scalability and flexibility of the continual learning model~\cite{zhao2021continual}. 
Pu \textit{et al.} construct a knowledge graph to represent the relationships and maintain the previous learned knowledge~\cite{pu2021lifelong}. 
Huang \textit{et al.} extend the problem into the unsupervised domain adaptation setting and propose a coordinated data replay mechanism with relational consistency restriction~\cite{huang2022lifelong}. 
Contrastive loss is introduced in~\cite{chen2022unsupervised} for sample rehearsal to provide better supervision. 
Ge \textit{et al.} transform the features of the new task into pseudo features of old tasks~\cite{ge2022lifelong}. 
Sun \textit{et al.} select patches instead of whole images for old data rehearsal, which provides more effective knowledge distillation~\cite{sun2022patch}. 
Although data rehearsal is effective for remembering past knowledge, it also brings the concern on data privacy issues. 
In this work, we propose to conduct color style transfer on the images of new tasks and implement a data-free continual UDA pipeline for person Re-ID. 

\subsection{Color style transfer}
Color style transfer is a practical technique for simulating knowledge from different domains. 
It is first proposed by~\cite{reinhard2001color} to match the mean and variance between source and target images. 
There are also works employing histograms of filter responses~\cite{pitie2005n}, illumination and texture~\cite{welsh2002transferring} as the transfer guidance. 
Recently, deep learning is incorporated into the color transfer process to provide finer conversion. 
Li \textit{et al.} employ an extra smoothing step to make the transferred images more realistic~\cite{li2018closed}. 
Yoo \textit{et al.} propose a wavelet corrected transfer to reduce information loss during pooling~\cite{yoo2019photorealistic}. 
An \textit{et al.} search for a more efficient network for color transfer~\cite{an2020ultrafast}. 
Chiu \textit{et al.} creates skip connections for high-frequency residuals~\cite{chiu2022photowct2}.

However, the involvement of neural networks also introduces longer calculation time. 
Besides, extra segmentation maps are often required to align the semantic information between images. 
In this work, we aim at conducting the color style transfer with as little extra computational cost as possible. 
Therefore, we adopt the training-free transfer with mean and variance as the main guidance. 
Based on that we design specific distribution capture and object-agnostic color style transfer modules for person Re-ID tasks. 

\section{Method}
In this section, we introduce the proposed Color Prompting method for data-free continual UDA person Re-ID. 
We first begin in Sec.~\ref{sec:color-bias} with the analysis on the color distribution gap in person Re-ID tasks. Sec.~\ref{sec:dist-capture} and Sec.~\ref{sec:agnostic-trans} then describe the design of our proposed distribution capture and color style transfer methods for simulating the source domain data. We further provide the detailed pipeline for the data-free continual learning scheme in Sec.~\ref{sec:continual-pipeline} and extend it to the domain generalizable Re-ID setting in Sec.~\ref{sec:extension-dg}. 

\begin{figure*}
\centering
    \includegraphics[width=\linewidth]{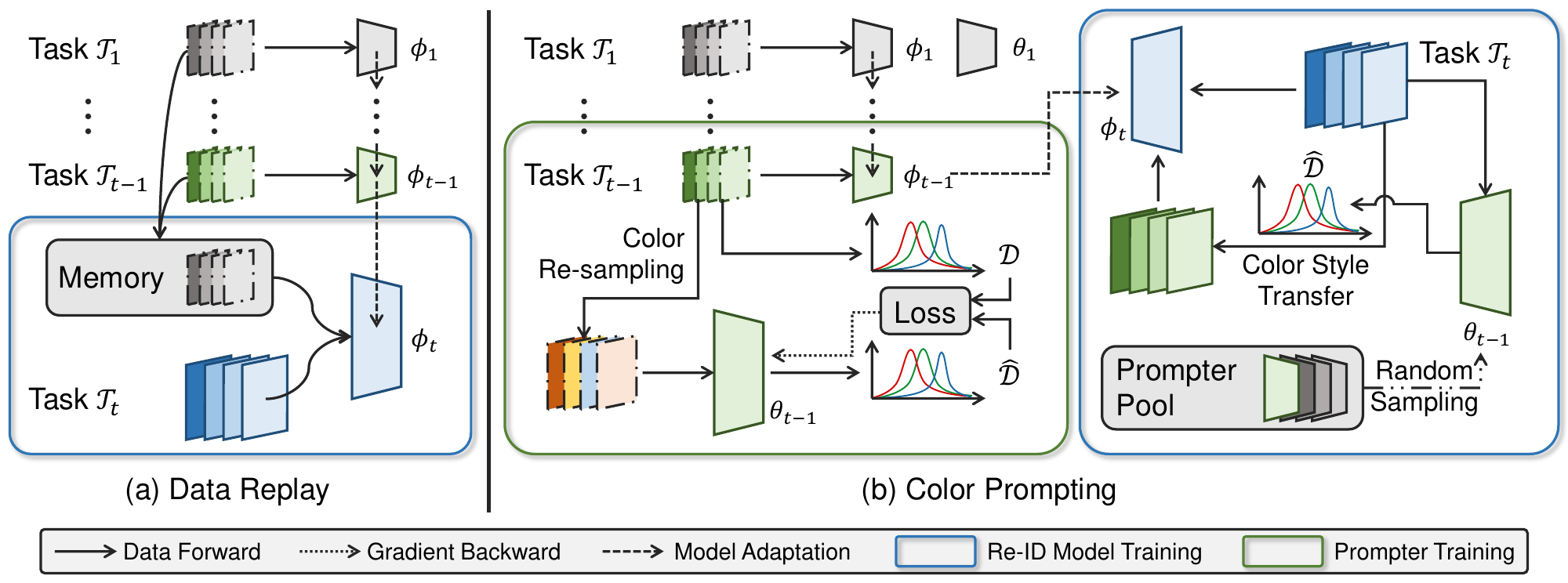}
    \caption{The training pipeline of (a) the data replay method and (b) our proposed Color Prompting method. Data replay maintains an image memory storing the samples of previous tasks. It provides direct information rehearsal, but has the risk of privacy leakage. The proposed Color Prompting (CoP) method distills the color distribution information into prompter networks. Then the information is served as color style transfer guidance for better anti-forgetting effects and training data diversity. }
    \label{fig:pipeline}
\end{figure*}

\subsection{Color Distribution Gap}
\label{sec:color-bias}
We first analyze the color distribution gap existing in the person Re-ID tasks. 
Re-ID systems involve a large number of cameras situated in diverse circumstances. 
Due to the long-term operation, maintenance, potential substitution and environmental differences, the images captured by different cameras can vary a lot in the resolution, pose, illumination and occlusion conditions. 
As a result, the appearance of the same identity can also have a large discrepancy across cameras within a single system, and even more so when considering images captured across different systems. 

Statistically, we summarize the color distribution of images captures by different cameras within the same system (intra-domain) and by different systems (inter-domain) in Fig.~\ref{fig:hist-intra} and Fig.~\ref{fig:hist-inter}, respectively. 
It is evident that color distribution gaps exist between images captured by different cameras, even in the same data domain. 
As shown in Fig.~\ref{fig:example-intra}, for the intra-domain scenario on Market-1501, the images captured by camera 5 have larger variance in illumination compared with those captured by camera 4, as also reflected in the distribution figures. 
The distribution gap poses a challenge to retrieve images across cameras. 
Fig.~\ref{fig:example-inter} showcases the color distribution gap across different data domains, representing a more challenging scenario.
When incrementally training the system during application, aiming for high performance on new tasks, the model can easily forget past knowledge due to the substantial distribution gap between data domains. 

Although the color distribution gap hinders the model generalization across domains, we also observe some unique characteristics in it. 
Specifically, the color distribution is distinguishable between different cameras or data domains, while remains consistent within each domain. 

Based on the aforementioned observations, we propose to address the forgetting issue during the continual learning process from the perspective of the color distribution gap, without requiring direct access to the data. 
Specifically, during the training stage of each task, we capture and store the color distribution information into a prompter neural network. 
When new tasks are introduced, the information is revisited and utilized to guide the color style transfer. 
By generating images with the color distribution of specific tasks, we effectively rehearse and retain the corresponding knowledge to a large extent. 
With limited risk of privacy leakage, the proposed Color Prompting (CoP) method significantly mitigates the catastrophic forgetting problem and enhance the robustness of the continual UDA Re-ID task. 

\subsection{Distribution Capture}
\label{sec:dist-capture}
For simulating images of specific domains, we need to first capture the corresponding color distributions. 
Since the color distribution gap is primarily induced by camera differences, a simple approach is to summarize the colors of images belonging to each camera. 
This summary is conducted during the training stage of each task.
The stored color distribution information is then revisited during incoming tasks to guide color style transfer. 
We refer to this straightforward solution as Camera Summary (CaS). 
Despite its ease of implementation and minimal computational requirements, CaS has the following drawbacks.
Firstly, CaS relies on camera labels for information summary. 
Thus, it is unsuitable for datasets where camera labels are not directly provided. 
Secondly, as the color distribution is summarized at the camera level, the diversity of obtained color distributions is limited by the number of cameras in the dataset. 

To address the above limitations, we propose to train a light-weighted prompter neural network to fit the color distribution of given data domains. 
Given an image as input, the prompter network is supposed to predict a reasonable color distribution that facilitates transferring the image to specific domains, conditioned on the image contents. 
The predicted color distribution then serves as guidance for the subsequent color style transfer process, enabling more flexible and domain-specific color adaptation. 

Since no previous data is explicitly stored, at any training stage, only images of the current task are available. 
Therefore, we first conduct random color style transfer inside the mini-batch to simulate images from different domains. 
The process is implemented by re-sampling from the color distribution of the current mini-batch, denoted as Color Re-sampling. 
Subsequently, the prompter is employed to predict the potential color distribution based on the content of the images. 
We explicitly restrict the predicted color distributions to be close to the original ones as,
\begin{equation}
\label{eq:prompter}
    \theta^{*}=\argmax_{\theta}\frac{1}{N}\sum^{N}_{i=1}\left(\mathcal{D}-\hat{\mathcal{D}}\right)^{2},
\end{equation}
where $\theta$ is the parameter of the prompter network, $N$ is the mini-batch size, $\mathcal{D}$ is the original color distribution and $\hat{\mathcal{D}}$ is the prediction of the prompter. 
Empirically, we summarize the mean $\mu$ and variance $\sigma$ values of the image color to represent the color distribution information. 

By fitting the color distribution conditioned on image contents, the prompter network is capable to generate reasonable color style transfer guidance for the incoming new tasks. 
Compared with CaS, where only a few groups of distributions are recorded, employing a prompter network significantly improves the diversity of generated color distributions. 
As the model fits the color distribution of the whole dataset, the generated distributions better reflect the characteristics of each image, resulting in more appropriate mean and variance values. 
Additionally, the prompter network does not require camera label as auxiliary prior knowledge, making it more practical for various Re-ID datasets. 



\begin{figure}[t]
    \centering
    \includegraphics[width=\columnwidth]{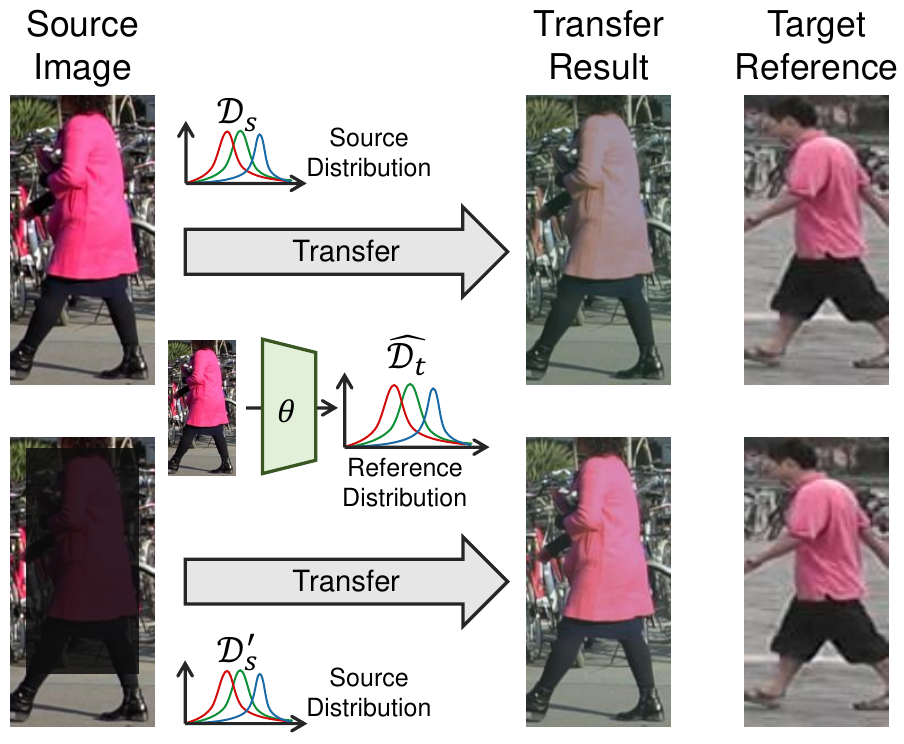}
    \caption{The color transfer example from MSMT-17 to Market-1501. The first row is plain transfer where all pixels are considered for source color distribution calculation. And the second row shows the result of our object-agnostic transfer. }
    \label{fig:example-frame}
\end{figure}

\subsection{Object-Agnostic color style transfer}
\label{sec:agnostic-trans}
After capturing the color distribution, we can conduct color style transfer for the images of new tasks to make them resemble the appearances of previous tasks. 
We mainly adopt the color style transfer method proposed in~\cite{reinhard2001color}. 
The process involves the following steps:
1) The images are first converted from the RGB color space to the LAB space, as the LAB space has the characteristic of independence between color channels. 
2) Then the images are normalized with the original mean and variance of each color channel.
3) The normalized images are transferred to specific color styles based on target mean and variance values. 
4) Finally the images are converted back to RGB color space for further training or inference. 

However, directly transferring the images without additional restrictions can lead to bias caused by the dominant color in the image. 
For example in Fig.~\ref{fig:example-frame}, if all the pixels are considered in the transferring process, the generated image may become biased towards green to balance the dominant red color in the original image. 
The dominance in color often arises from the clothing of pedestrians in the images. 

Therefore, we propose to design an object-agnostic color style transfer module to avoid such bias. 
Since Re-ID datasets have typically included a detection process, the objects are mostly distributed in the central area of the images. 
We leverage the property by cropping out the central area when calculating the mean and variance of the target domain images to be transferred. 
By considering only the frame area of the images, the impact of dominant color distribution gap can be significantly reduced, as shown in Fig.~\ref{fig:example-frame}. 
And the transferred image is more similar to the target style. 
As the prompter network is trained on a large amount of images under each camera, the dominant color bias is already mitigated, for which the predicted mean and variance values from the prompter can directly serve as transfer guidance. 



\subsection{Data-free Continual Learning}
\label{sec:continual-pipeline}
With the above two proposed components, we can transfer images from any domains to have specific color distributions. 
Below we illustrate the detailed continual UDA person Re-ID pipeline with the proposed data-free CoP method. 

At the training stage $\mathcal{T}_{i}$, alongside training the Re-ID models for the current task, a prompter network $\theta_{i}$ is also trained to capture the corresponding color distribution. 
Thereby we obtain a prompter pool $\mathcal{P}_c$ containing the color distribution information of all the previous tasks. 
Then at each training epoch of the incoming new task $\mathcal{T}_{i+1}$, we randomly sample a prompter network from the pool $\theta_{j}\in\mathcal{P}_c\:(1\leq j\leq i)$.
The selected prompter is used to provide information rehearsal of the corresponding domain. 

We optimize the Re-ID model with the cluster-level contrastive loss~\cite{dai2022cluster,ge2020self}.
A prototype memory $\mathcal{C}$ is maintained containing the prototypes of all the classes. 
After each clustering process, $\mathcal{C}$ is re-initialized with the average feature of the clusters. 
Given a embedded feature $f$, the contrastive loss can be formulated as,
\begin{equation}
    \mathcal{L}(f)=-\log\frac{\exp\left(\langle f,c^{+}\rangle/\tau\right)}{\sum^{N_c}_{i=0}\exp\left(\langle f,c_i\rangle/\tau\right)},
\end{equation}
where $c_i$ represents the prototype embedding for the $i$-th class, $c^+$ indicates the prototype with the same label as the feature $f$, and $N_c$ is the total class number. $\langle\cdot,\cdot\rangle$ stands for the inner product operation, and $\tau$ is a temperature hyper-parameter, which is empirically set as 0.05. 
And the prototype $c_i$ is updated after each time a sample with the corresponding label is involved into training as,
\begin{equation}
    c_i\leftarrow \alpha c_i + (1-\alpha)\cdot\frac{1}{|\mathcal{B}_i|}\sum_{f_k\in \mathcal{B}_i}f_k,
\end{equation}
where $\mathcal{B}_i$ represents the sample set inside the current mini-batch with the identity label $i$ and $\alpha$ is a momentum hyper-parameter, which is empirically set as 0.2. 

Adding the transferred data into the training updates the whole training objective into,
\begin{equation}
\label{eq:loss}
    \mathcal{L}=\sum^{N}_{i=1}\lambda\mathcal{L}(f_i)+(1-\lambda)\mathcal{L}(\hat{f}_{i}),
\end{equation}
where the $\hat{f}_{i}$ refers to the extracted features of transferred images, and $\lambda$ is a weighting hyper-parameter. 
Along with the loss calculation, the prototype memory $C$ is also updated with $\hat{f}_i$. 

The detailed training pipeline is illustrated in Fig.~\ref{fig:pipeline}. 
Data replay methods maintain an image memory storing the samples of past tasks, allowing for direct information rehearsal. 
However, such scheme restricts the diversity of the rehearsal information, and carries the risk of privacy leakage. 
On the contrary, our proposed Color Prompting (CoP) method stores the color distribution in the prompter networks, which in turn guide the color style transfer for information rehearsal of past tasks. 
As a result, the actual images of past tasks are not directly required. However, CoP allows for much more diverse rehearsal information compared with data replay. 

\definecolor{Gray}{gray}{0.9}
\begin{table*}[t]
\centering
\caption{Seen-domain results of the continual unsupervised domain adaptation experiment. The experiment is conducted on the Market-1501 $\to$ CUHK-SYSU $\to$ MSMT-17 task. MA: Market-1501; SY: CUHK-SYSU; MS: MSMT-17. }
\label{tab:continual-seen}
\small
\setlength{\tabcolsep}{7.5pt}
\begin{tabular}{c|cccccc|cc}
    \toprule
    \multirow{3}{*}{Methods} & \multicolumn{2}{c}{MA~\cite{zheng2015scalable}} & \multicolumn{2}{c}{SY~\cite{xiao2016end}} & \multicolumn{2}{c|}{MS~\cite{wei2018person}} & \multicolumn{2}{c}{Avg} \\
     & mAP & R-1 & mAP & R-1 & mAP & R-1 & mAP & R-1 \\
    \midrule
    Baseline & 29.6 & 55.7 & 75.4 & 78.7 & 23.8 & 48.6 & 42.9 & 61.0 \\
    \rowcolor{Gray}
    Replay & 35.8 & 62.0 & \textbf{78.5} & \textbf{81.9} & 24.0 & 48.5 & 46.1 \textcolor{purple}{(+3.2)} & 64.1 \textcolor{purple}{(+3.1)} \\
    CoP & \textbf{43.1} & \textbf{73.0} & 75.2 & 77.9 & \textbf{30.8} & \textbf{61.5} & \textbf{49.7} \textcolor{purple}{(+6.8)} & \textbf{70.8} \textcolor{purple}{(+9.8)} \\
    \bottomrule
\end{tabular}
\end{table*}

\begin{table*}[t]
\centering
\caption{Unseen-domain results of the continual unsupervised domain adaptation experiment. The experiment is conducted on the Market-1501 $\to$ CUHK-SYSU $\to$ MSMT-17 task.}
\label{tab:continual-unseen}
\small
\setlength{\tabcolsep}{6.9pt}
\begin{tabular}{c|cccccccc|cc}
    \toprule
    \multirow{2}{*}{Methods} & \multicolumn{2}{c}{CUHK02~\cite{li2013locally}} & \multicolumn{2}{c}{CUHK03~\cite{li2014deepreid}} & \multicolumn{2}{c}{GRID~\cite{loy2009multi}} & \multicolumn{2}{c|}{iLIDS~\cite{wang2014person}} & \multicolumn{2}{c}{Avg} \\
     & mAP & R-1 & mAP & R-1 & mAP & R-1 & mAP & R-1 & mAP & R-1 \\
    \midrule
    Baseline & 53.0 & 50.0 & 21.9 & 20.8 & 19.5 & 13.6 & 65.4 & 54.3 & 40.0 & 34.7 \\
    \rowcolor{Gray}
    Replay & 53.4 & 50.4 & 21.7 & 20.2 & 22.3 & 14.4 & 68.1 & 58.0 & 41.4 \textcolor{purple}{(+1.4)} & 35.8 \textcolor{purple}{(+1.1)} \\
    CoP & \textbf{61.5} & \textbf{60.9} & \textbf{29.9} & \textbf{31.6} & \textbf{30.6} & \textbf{20.0} & \textbf{71.5} & \textbf{63.0} & \textbf{48.4} \textcolor{purple}{(+8.4)} & \textbf{43.9} \textcolor{purple}{(+9.2)} \\
    \bottomrule
\end{tabular}
\end{table*}

\subsection{Extension to Generalizable Re-Identification}
\label{sec:extension-dg}
In addition to the standard continual UDA Re-ID scenario, the proposed CoP method is also capable to enhance the generalization capability of any Re-ID models. 
The effectiveness of CoP on the model generalization can be illustrated by two folds. 

Firstly, in scenarios where specific domain information is not provided, the Color Prompting can be considered as a data augmentation method. 
During the source domain pre-training process, for example, we calculate the mean and variation for each image inside a mini-batch and randomly shuffle the distributions. 
The shuffled distributions are then used as the color style transfer targets. 
In comparison to other methods that generate images with different camera styles~\cite{zhong2018camstyle}, CoP requires no extra training consumption, and is more efficient at the inference stage. 
We denote the above operation as Color Shuffling.

Secondly, in scenarios where only a small amount of data is collected for new domains, CoP is capable of generating a large number of images with the corresponding styles.  
Compared with previous methods based on generative adversarial networks that require target domain data for training~\cite{deng2018image,wei2018person}, CoP is a more flexible solution to for real-world Re-ID systems with limited data collection in new scenarios. 

\section{Experimental Results}
\subsection{Datasets and Evaluation Metrics}
We conduct experiments on multiple popular person Re-ID datasets, including:

\textbf{Market-1501}~\cite{zheng2015scalable} contains 32,668 images of 1,501 identities captured by 6 different cameras. 

\textbf{MSMT-17}~\cite{wei2018person} is a large-scale Re-ID dataset, consisted of 126,441 images from 4,101 identities. 

\textbf{CUHK-SYSU}~\cite{xiao2016end} is composed of 18,184 images of 8,432 identities from diverse scenarios. 

\textbf{CUHK02}~\cite{li2013locally} has 7,264 manually cropped images of 1816 identities. 

\textbf{CUHK03}~\cite{li2014deepreid} contains 13,164 images that are manually cropped and automatically detected form 1,360 identities. 



\textbf{GRID}~\cite{loy2009multi} contains 250 pedestrian image pairs from different camera views. 

\textbf{iLIDS}~\cite{wang2014person} is consisted with images from 300 persons with challenging environments. 

Evaluation Metrics include rank-1 accuracy from Cumulative Matching Characteristic (CMC) curve and mean average precision (mAP).

\subsection{Experiment Settings}
We evaluate our proposed method on a continual unsupervised domain adaptive person Re-ID training pipeline. 
The model is first trained in a supervised manner on Market-1501 to obtain basic knowledge on the task. 
The pre-training step serves as a starting point for the continual UDA process. 
The model is then incrementally trained on the CUHK-SYSU and MSMT-17 datasets with no annotation information provided. 
After the whole training process, the model is evaluated on all the seen and unseen datasets to assess its performance and generalization capabilities. 

\subsection{Implementation Details}
Following the common settings adopted in Re-ID tasks, we employ ResNet50~\cite{he2016deep} pre-trained on ImageNet~\cite{deng2009imagenet} as the backbone network. 
For the source domain training, the model is trained for 80 epochs, using an Adam optimizer~\cite{kingma2014adam} with a learning rate of 3.5e-4. 
Then for domain adaptation training, the model is trained for 60 epochs on each dataset. 
Adam is adopted as well, with a learning rate of 2e-4. 
The mini-batch size is set as 64 and 128 for the source supervised training and unsupervised domain adaptation, respectively. 
The images are resized to 256$\times$128.
Random flipping, random cropping and random erasing are employed as augmentation. 

For the prompter network, ShuffleNet-V2 is adopted as the backbone architecture~\cite{ma2018shufflenet}. 
An Adam optimizer with a learning rate of 2e-4 is adopted for optimizing. 
The loss weight $\lambda$ in Eq.~\ref{eq:loss} is set as 0.5. 
All the experiments are conducted on a single Tesla-A100 GPU. 

\begin{table}[t]
    \centering
    \caption{Experiment results of the domain generalization tasks. The model is trained on Market-1501 in a supervised manner. We report the performance on CUHK-SYSU and MSMT-17. }
    \label{tab:generalization}
    \setlength{\tabcolsep}{11pt}
    \begin{tabular}{c|cccc}
        \toprule
        \multirow{2}{*}{Method} & \multicolumn{2}{c}{SY~\cite{xiao2016end}} & \multicolumn{2}{c}{MS~\cite{wei2018person}} \\
         & mAP & R-1 & mAP & R-1 \\
        \midrule
        Baseline & 69.9 & 73.4 & 2.6 & 7.9 \\
        \rowcolor{Gray}
        Joint & 64.3 & 68.9 & 3.0 & 9.2 \\
        CoP & \textbf{75.0} & \textbf{77.5} & \textbf{8.2} & \textbf{24.6} \\
        \bottomrule
    \end{tabular}
\end{table}

\subsection{Continual Learning Experiments}
In the main continual learning experiments, we compare our proposed Color Prompting (CoP) method with baseline and replay methods. 
The baseline method refers to naive domain adaptation without any specific design for anti-forgetting. 
For the replay method, we maintain an image memory containing 512 images and update it with a reservoir strategy after each task~\cite{chaudhry2019tiny}. 
The replay data is then jointly trained with the current task. 

The performance on each dataset after training on CUHK-SYSU and MSMT-17 is presented in Tab.~\ref{tab:continual-seen}. 
Our proposed CoP method achieves the highest performance on Market-1501 and MSMT-17 and comparable performance on CUHK-SYSU. 
As Market-1501 and MSMT-17 are the first and last tasks, respectively, the experiment results suggest that CoP improves both the adaptation and anti-forgetting capabilities by a large margin. 
Notably, CoP surpasses the replay method on these two datasets, while not requiring the original data. 
It demonstrates that color style transfer alone provides abundant information for retaining the knowledge of past tasks. 

Additionally, we evaluate the trained model on unseen datasets after the entire training process in Tab.~\ref{tab:continual-unseen} to validate the generalization capability of our proposed CoP method. 
The replay method, which introduces extra samples for training, only achieves a slight performance improvement compared with the baseline. 
Comparatively, CoP surpasses both the baseline and replay methods by a large margin.
The experiment results validate that CoP substantially increases the data diversity, leading to significant generalization improvement. 

\begin{table}[t]
    \centering
    \caption{Component analysis experiment results for the proposed CoP method. The experiment is conducted on the Market-1501 $\to$ MSMT-17 task. PR: color style transfer at the source domain pre-training stage; CT: color style tranfer for domain adaptation; PN: prompter network employed for mean and variation generation; OA: object-agnostic color style transfer. }
    \label{tab:comp-ana}
    \setlength{\tabcolsep}{5.6pt}
    \begin{tabular}{cccc|cccc}
        \toprule
        \multicolumn{4}{c|}{Components} & \multicolumn{2}{c}{MA~\cite{zheng2015scalable}} & \multicolumn{2}{c}{MS~\cite{wei2018person}} \\
        PR & CT & PN & OA & mAP & R-1 & mAP & R-1 \\
        \midrule
         & & & & 31.3 & 58.2 & 24.6 & 49.9 \\
         \rowcolor{Gray}
         & \checkmark & & & 38.2 & 69.2 & 26.4 & 55.9 \\
         & \checkmark & \checkmark & & 38.9 & 70.1 & 27.3 & 57.3 \\
         \rowcolor{Gray}
         & \checkmark & & \checkmark & 39.2 & 70.3 & 27.8 & 57.9 \\
         & \checkmark & \checkmark & \checkmark & \textbf{39.4} & \textbf{71.3} & \textbf{29.7} & \textbf{59.8} \\
        \midrule
        \checkmark & & & & 39.0 & 67.0 & \textbf{33.2} & 61.7 \\
         \rowcolor{Gray}
        \checkmark & \checkmark & & & 41.0 & 72.0 & 29.9 & 60.9 \\
        \checkmark & \checkmark & \checkmark & & 40.9 & 72.2 & 30.1 & 61.3 \\
         \rowcolor{Gray}
        \checkmark & \checkmark & & \checkmark & 42.4 & 74.1 & 31.3 & 62.7 \\
        \checkmark & \checkmark & \checkmark & \checkmark & \textbf{42.2} & \textbf{73.1} & 32.5 & \textbf{63.4} \\
        \bottomrule
    \end{tabular}
\end{table}

\subsection{Generalization Experiments}
In addition to the continual learning setting, CoP can also be applied to domain generalization scenarios. 
We evaluate the effectiveness of CoP in Tab.~\ref{tab:generalization}. 
Specifically, we train the model on Market-1501 in a supervised manner as the baseline. 
Then we randomly select 16 classes from CUHK-SYSU or MSMT-17 datasets, each with 8 images, to simulate a fast annotation scenario.
The images are directly added into the training process to provide information of the new domain, denoted as ``Joint'' in Tab.~\ref{tab:generalization}. 
Finally for CoP, the samples are regard as color style transfer references, without label information involved. 

\begin{figure*}[t]
    \centering
    \includegraphics[width=\linewidth]{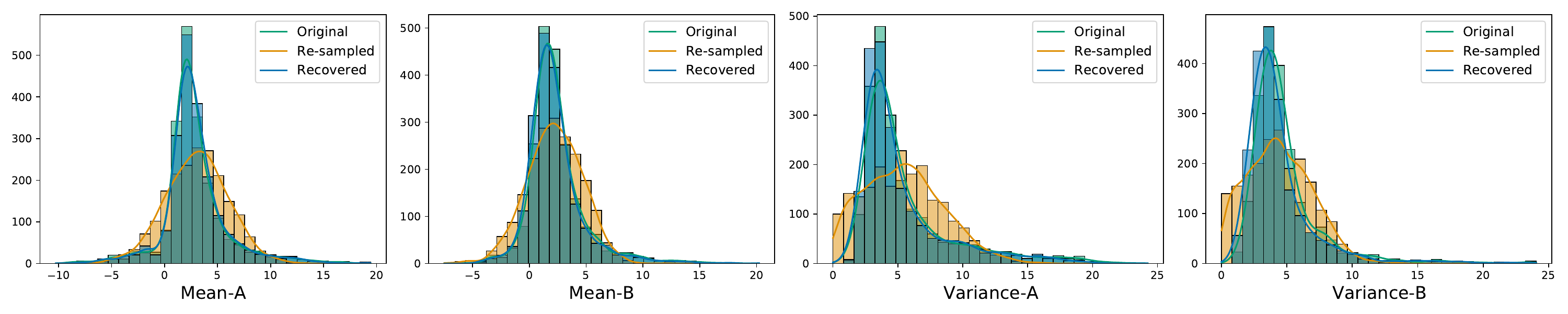}
    \caption{The color distribution of original, re-sampled and recovered images captured by camera 0 on Market-1501. The prompter network accurately recovers the color distribution. }
    \label{fig:dist-recover}
\end{figure*}

The small amount of target domain samples slightly improves the performance on MSMT-17, but degrades the performance on CUHK-SYSU by a large margin. 
In comparison, the model trained with CoP achieves a significant boost in the generalization performance. 
It is noticeable that the performance on CUHK-SYSU is quite close to the UDA results without utilizing the samples for direct supervision. 
The experiment results validate that the proposed CoP method is capable of fast adaptation on new unseen domains, even without the need for annotations. 
The characteristic makes CoP a powerful and practical solution for real-world Re-ID systems. 

\begin{table}[t]
    \centering
    \caption{Experiment results when no camera label is provided. The experiment is conducted on the CUHK-SYSU $\to$ MSMT-17 task. By default object-agnostic color style transfer is applied. }
    \label{tab:comp-ana-cuhk}
    \setlength{\tabcolsep}{10.4pt}
    \begin{tabular}{c|cccc}
        \toprule
        \multirow{2}{*}{Method} & \multicolumn{2}{c}{SY~\cite{xiao2016end}} & \multicolumn{2}{c}{MS~\cite{wei2018person}} \\
         & mAP & R-1 & mAP & R-1 \\
        \midrule
        CaS & 72.5 & 74.7 & 29.6 & 59.8 \\
         \rowcolor{Gray}
        Prompter & \textbf{74.8} & \textbf{78.0} & \textbf{31.4} & \textbf{61.7} \\
        \bottomrule
    \end{tabular}
\end{table}

\begin{table}[t]
    \centering
    \caption{Effectiveness validation of the color style transfer towards specific domains. The experiment is conducted on the Market-1501 $\to$ MSMT-17 task. For fair comparison, random color style transfer is included for all the models at the source domain pre-training stage. }
    \label{tab:domain-spec}
    \setlength{\tabcolsep}{11pt}
    \begin{tabular}{c|cccc}
        \toprule
        \multirow{2}{*}{Method} & \multicolumn{2}{c}{MA~\cite{zheng2015scalable}} & \multicolumn{2}{c}{MS~\cite{wei2018person}} \\
         & mAP & R-1 & mAP & R-1 \\
        \midrule
        Baseline & 39.0 & 67.0 & \textbf{33.2} & 61.7 \\
         \rowcolor{Gray}
        Shuffle & 40.7 & 71.7 & 32.1 & \textbf{64.0} \\
        CoP & \textbf{42.2} & \textbf{73.1} & 32.5 & 63.4 \\
        \bottomrule
    \end{tabular}
\end{table}

\section{Ablation Study and Analysis}
\subsection{Component Analysis}
We conduct an extensive analysis on the effectiveness of each component by evaluating the performance of various combinations in Tab.~\ref{tab:comp-ana}. 
The experiments are conducted in two groups. 
In the upper part we present the results without adding the Color Shuffling augmentation at the source domain pre-training stage, and the lower part presents the results with the augmentation. 

In the upper part, adding the color style transfer to the baseline model results in a substantial improvement in the anti-forgetting effects. 
For color transfer without prompter network, the Camera Summary (CaS) module is adopted. 
Substituting the CaS with the prompter network and adding object-agnostic color style transfer both receive slight performance improvement. 
Combining all the components, the CoP achieves the highest performance on all metrics, especially with a remarkable 13.1\% boost on the source domain rank-1 metric, demonstrating the extraordinary anti-forgetting effects. 

In the lower part, the Color Shuffling augmentation largely increases the sample diversity, which improves both the target domain performance and anti-forgetting effects during the adaptation training process. 
Based on the high baseline, adding the color style transfer towards past domains further reduces the performance degradation on the source domain during the adaptation, especially on the rank-1 metric. 
While the target domain performance drops when adding the style transfer alone, the complete CoP method still achieves the highest rank-1 on MSMT-17. 

\begin{figure}[t]
    \centering
    \includegraphics[width=\columnwidth]{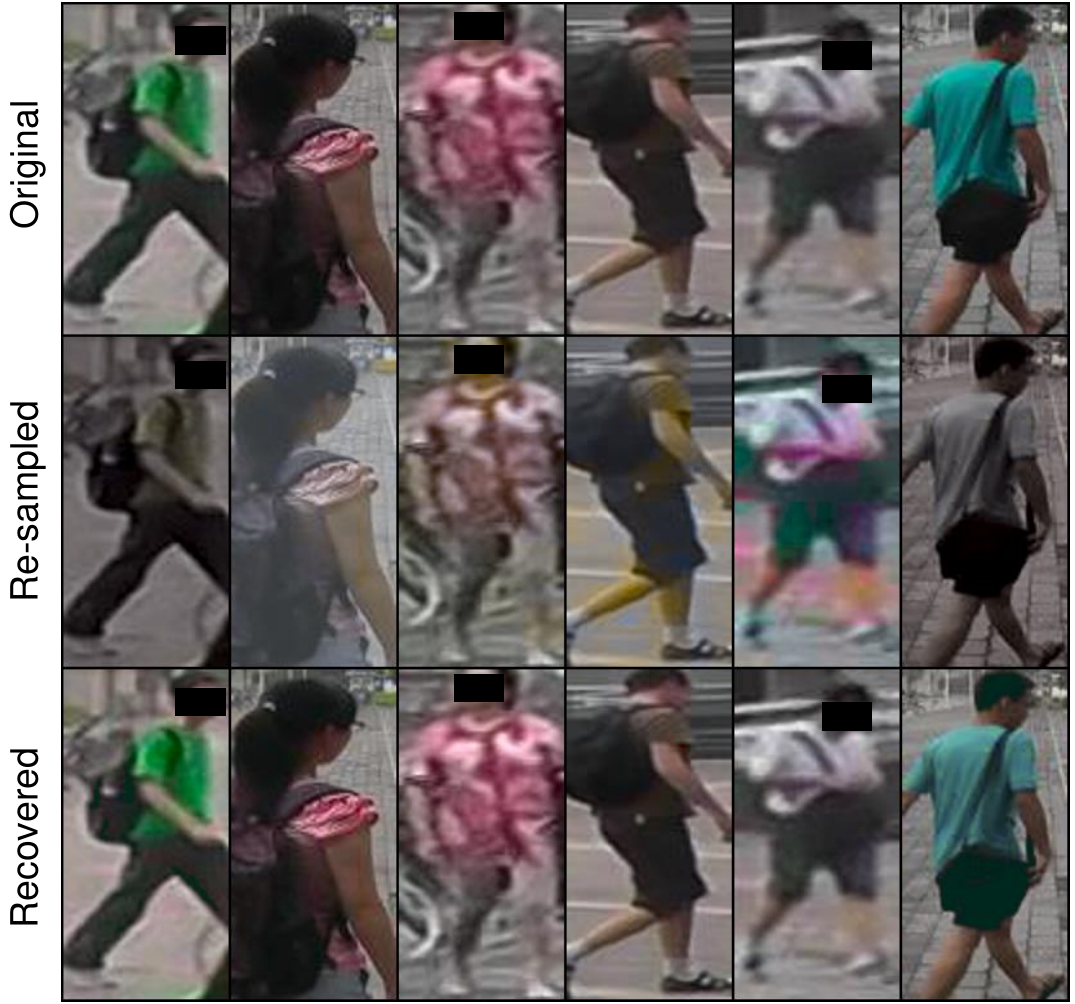}
    \caption{Example images of the color distribution recover experiment. The three rows show the original images, the images transferred through distribution re-sampling, and the recovered images guided by prompter-predicted distributions. }
    \label{fig:example-trans}
\end{figure}

In order to further validate the rationality of the prompter network, we conduct the ablation study on the CUHK-SYSU $\to$ MSMT17 task in Tab.~\ref{tab:comp-ana-cuhk}. 
Since no camera labels are provided in CUHK-SYSU, CaS (without prompter network) can only provide a single group of mean and variance as the color style transfer guidance. 
As a result, the improvement of color style transfer is limited due to the poor diversity. 
Comparatively, substituting CaS with prompter network largely mitigates the diversity issue and fully exploits the effects of color style transfer. 

In addition, in Tab.~\ref{tab:domain-spec} we validate the effectiveness of CoP in simulating the information of specific domains. 
We denote the color style transfer guided by the shuffled color distribution inside the mini-batch as ``shuffle'', which is also adopted in the pre-training process to increase the generalization capability. 
The shuffle augmentation largely improves the sample diversity, and provides general performance boosts, especially on the target domain. 
However, there still exists a performance gap on the anti-forgetting effects on the source domain from the proposed CoP method. 
It validates that CoP provides more specific color style transfer guidance towards the source domain, and helps better reduce the knowledge forgetting during the continual training process. 

\begin{figure}[t]
    \centering
    \includegraphics[width=\columnwidth]{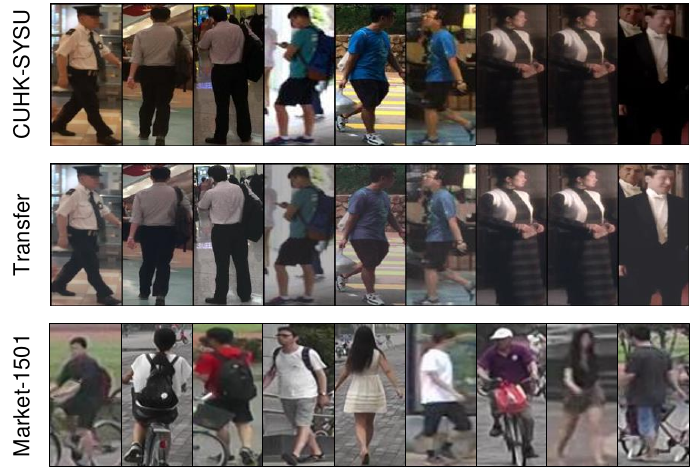}
    \caption{Example images of the color style transfer from CUHK-SYSU to Market-1501. }
    \label{fig:example-apply}
\end{figure}

\subsection{Visualization Analysis}
In this section, we provide some visualization results to support the effectiveness of our proposed CoP method. 
Initially, we show the effects of recovering the original color distribution in Fig.~\ref{fig:dist-recover}. 
The experiments are conducted on the images captured by camera 0 on Market-1501. 
We first conduct a re-sampling operation that is also adopted in the training process of the prompter network to generate out-of-distribution samples. 
Then the prompter is adopted to predict recovered mean and variance values based on the image contents. 
As shown, the re-sampled distribution has distinguishable differences from the original distribution, while the prompter is capable to recall the color for the images, and recover the correct distribution. 
It validates that the prompter successfully captures the color distribution of the training dataset. 

\begin{figure}[t]
\begin{subfigure}[t]{0.245\textwidth}
    \includegraphics[width=\textwidth]{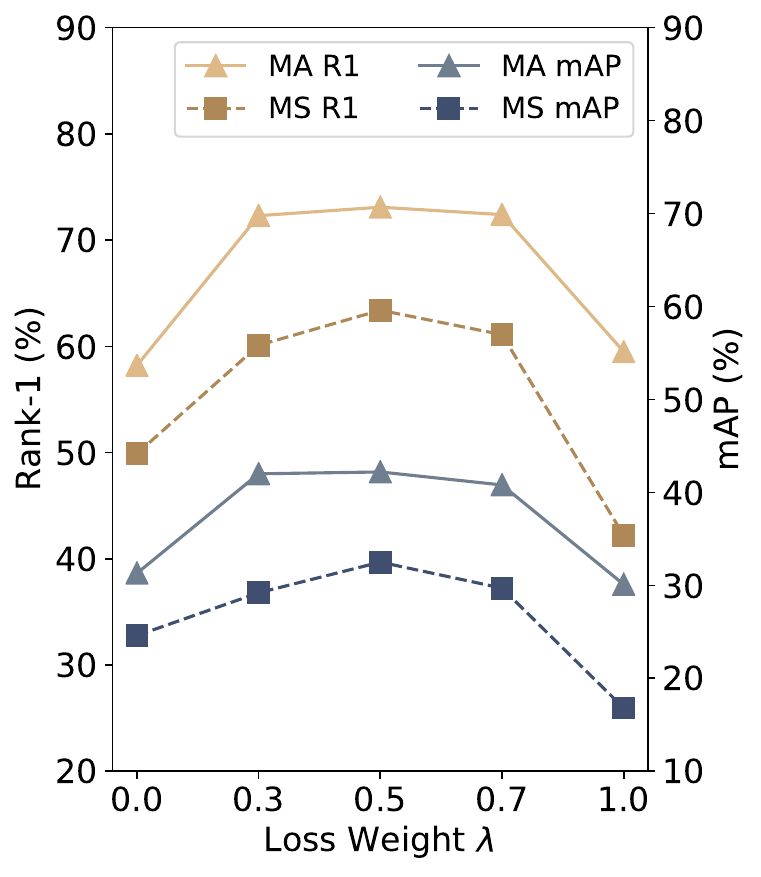}
    \caption{}
    \label{fig:lambda}
\end{subfigure}
\hfill
\begin{subfigure}[t]{0.225\textwidth}
    \includegraphics[width=\textwidth]{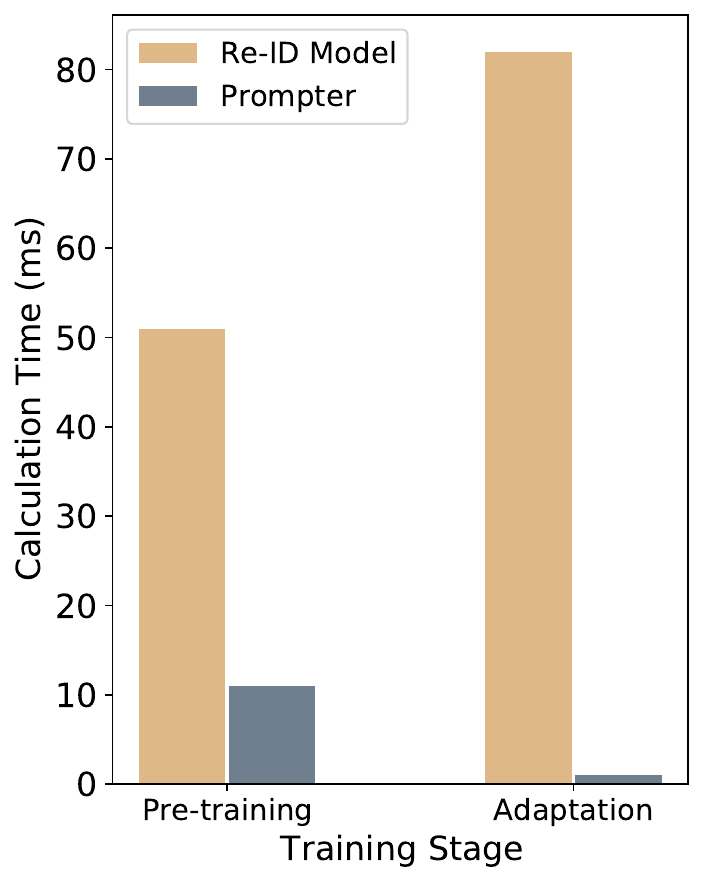}
    \caption{}
    \label{fig:cost}
\end{subfigure}
\caption{(a) The parameter analysis curves on the loss weight $\lambda$. The experiment is conducted on the Market-1501 $\to$ MSMT-17 task. (b) The calculation time comparison between the prompter and other computation processes. }
\end{figure}

\begin{figure*}
    \centering
    \includegraphics[width=0.9\linewidth]{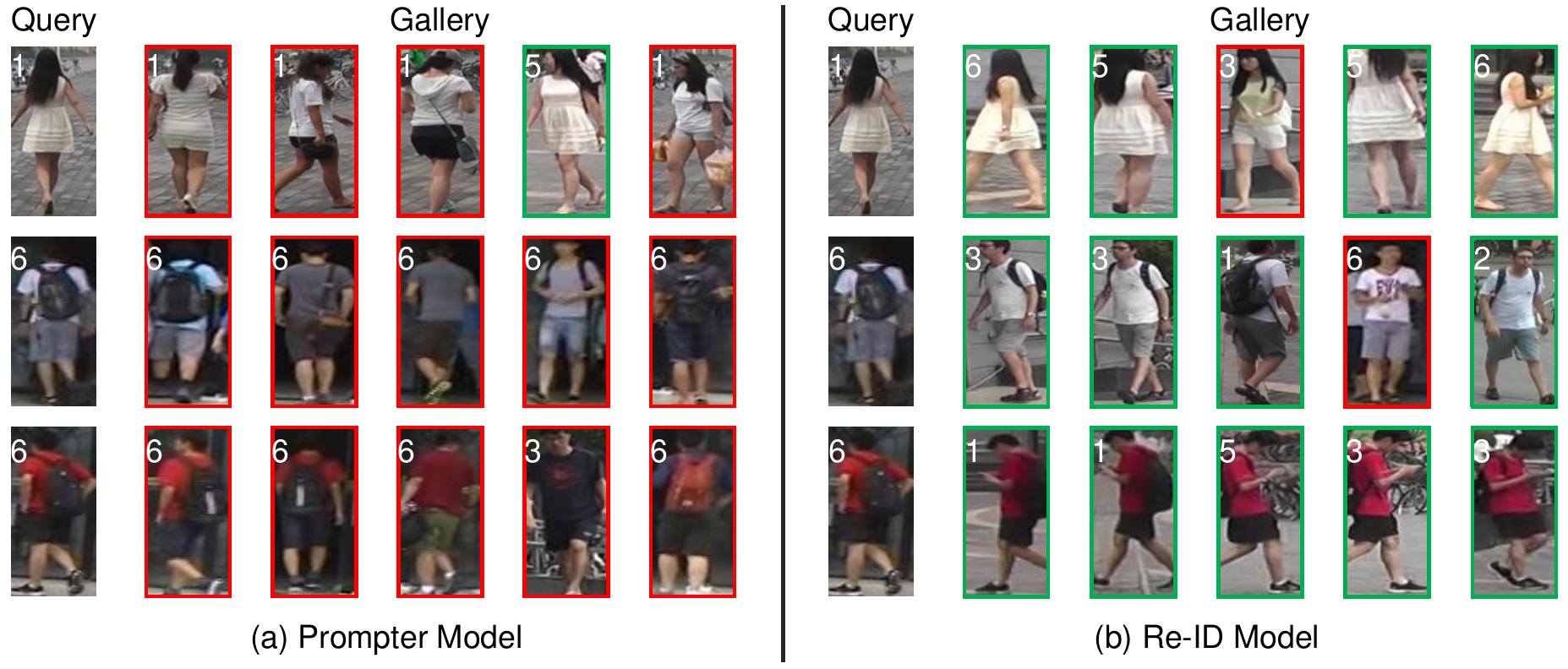}
    \caption{The retrieval list comparison between the prompter model and the Re-ID model. The  \textcolor{green}{green} and \textcolor{red}{red} boxes represent the positive and negative samples. On the upper-left of the image we attach the camera that captures the corresponding image. }
    \label{fig:privacy}
\end{figure*}

Secondly, we provide example images from the re-sampling experiment in Fig.~\ref{fig:example-trans}. 
The re-sampling operation introduces severe color distortion into the images, leading to hue shifts (column 4), desaturation (column 1-3, 6) and color disorder (column 5). 
In spite of the challenging color modifications, the prompter network accurately recovers the original color for these images, as shown in the last row of the figure. 

\begin{table}[t]
    \centering
    \caption{The performance comparison between the prompter network, ResNet-50 pre-trained on ImageNet and on Re-ID datasets. The prompter is trained on Market-1501 and CUHK-SYSU for the corresponding evaluation. }
    \label{tab:privacy}
    \setlength{\tabcolsep}{8pt}
    \begin{tabular}{c|cccc}
        \toprule
        \multirow{2}{*}{Method} & \multicolumn{2}{c}{MA~\cite{zheng2015scalable}} & \multicolumn{2}{c}{SY~\cite{xiao2016end}} \\
         & mAP & R-1 & mAP & R-1 \\
        \midrule
        Prompter-Raw & 4.7 & 16.1 & 34.8 & 39.3 \\
        Prompter-Trained & 4.4 & 13.7 & 38.5 & 42.4 \\
        ResNet-50-ReID & 86.0 & 94.1 & 89.9 & 91.4 \\
        \bottomrule
    \end{tabular}
\end{table}

Besides, the examples in the actual appliance are also shown in Fig.~\ref{fig:example-apply}, where the images of CUHK-SYSU are transferred to have the style of Market-1501. 
The images of CUHK-SYSU have higher saturation and contrast compared with those of Market-1501. 
By employing the prompter network, the color of transferred images is made more plain and similar to the color style of Market-1501, while still preserving the image details for more accurate Re-ID feature extraction. 
Notably, the prompter network never accesses the data of CUHK-SYSU during its training, yet it exhibits remarkable robustness in accurately predicting proper color distributions for the style transfer. 

\subsection{Parameter Analysis}
\textbf{Loss Weight $\lambda$. } The loss weight $\lambda$ controls the optimization preference between the Re-ID loss on the original data and the transferred data. 
We investigate the effect of $\lambda$ on the performance of CoP method in Fig.~\ref{fig:lambda}. 
When $\lambda$ is set to 0, the model is only supervised by the original data, which serves as the baseline in this work. 
On the other extreme, when $\lambda$ is set to 1, the model completely relies on the transferred data without the accurate supervision of original target domain data.
It leads to chaotic Re-ID training and even worse performance than baseline. 
The CoP generally brings significant improvements in terms of both anti-forgetting and target domain performance when $\lambda$ lies between 0.3 and 0.7. 
This suggests that a balanced combination of the original data and the transferred data is crucial for achieving the best performance. 
Balancing both aspects, we set $\lambda$ as 0.5 in our experiments. 

\subsection{Computational Cost}
We record the calculation time of the prompter network and the color style transfer process in each iteration in Fig.~\ref{fig:cost}. 
During pre-training, the prompter is trained to fit the color distribution, which requires approximately one fifth of the Re-ID model training time. 
For the adaptation process, where the prompter predicts the potential color distribution, the inference time of prompter and the color transfer process combined is negligible. 
The calculation time comparison suggests that the performance improvement brought by CoP is not achieved at the expense of large extra computational cost. 
On the contrary, CoP is not only effective, but also efficient in rehearsing the information of past tasks without imposing a heavy computational burden. 

\subsection{Privacy Protection}
The prompter network is proposed to avoid privacy leakage, for which we design a retrieval list comparison between the prompter and the trained Re-ID model in Fig.~\ref{fig:privacy}. 
For each row we compare the gallery images retrieved with the same query image. 
We exclude images with the same identity and camera label simultaneously as the query image, which is also adopted in Re-ID evaluation processes. 
On the right side, the Re-ID model focuses on detailed distinctions across different camera views and retrieve correct samples. 
Oppositely, the prompter model mainly focuses on the overall color distribution and background similarity. 
Notably, most of the top-similar samples retrieved by the prompter are captured by the same camera. 
It validates that the color distribution rather than identity information is learned by the prompter networks. 

Quantitatively, we compare the performance of the raw prompter network (backbone pre-trained on ImageNet~\cite{ma2018shufflenet}), the trained prompter network and a ResNet-50 trained on the Re-ID datasets in Tab.~\ref{tab:privacy}. 
Although having well captured the color distribution, the trained prompter only slightly surpasses the raw network on CUHK-SYSU, while performs worse on Market-1501.
Compared with the ResNet-50 trained on Re-ID datasets, the performance is far from practical appliance. 

From the analysis, we can confidently state that the features extracted by the prompter network are not related to specific identities, and therefore, the adoption of the prompter network involves limited risk of privacy leakage.

\subsection{Comparison with GAN}
There are also some previous works conducting style transfer with the Generative Adversarial Network (GAN)~\cite{wei2018person,deng2018image}. Here we explain the differences between GAN and our proposed CoP, and the reason why CoP is more practical. 

Firstly, the GAN training requires the data accessibility of both source and target domains, which does not meet the data-free scenario in this work. 
CoP trains the prompter network together with the standard Re-ID network for the current task, which does not rely on data of other domains. 
Secondly, the color transfer process of GAN typically involves down-sampling and up-sampling of images, which can lead to information loss and potentially degrade the image quality. 
In comparison, CoP only modify the color distribution of images, without altering the structural features in the images. 
Besides, identity information is often inadvertently captured from the source domain during the GAN training process for Re-ID. 
Oppositely, CoP focuses solely on the color distribution and is designed to be privacy-preserving by not capturing identity information. 

\section{Conclusion}
In this work, we propose a Color Prompting (CoP) method for the continual unsupervised domain adaptive person re-identification (Re-ID) task. 
It offers a data-free and privacy-friendly solution by effectively capturing and transferring color distribution information of previous tasks. 
CoP significantly improves the anti-forgetting effects, target domain performance, and generalization capability compared with data replay methods. 
The proposed method opens up new possibilities for practical and efficient continual learning in Re-ID systems without compromising privacy. 
It is our hope that this work will inspire further advancements in the field of continual Re-ID and contribute to the development of more robust and adaptable person Re-ID systems. 

\backmatter

\bmhead{Supplementary information}
The scripts are attached for reproduction in the supplementary material. 




\section*{Declarations}
\begin{itemize}
\item Availability of data and materials

The datasets adopted can be requested and downloaded through the following links:
\begin{itemize}
    \item \textbf{Market-1501} \href{https://zheng-lab.cecs.anu.edu.au/Project/project_reid.html}{link}
    \item \textbf{CUHK-SYSU} \href{http://www.ee.cuhk.edu.hk/~xgwang/PS/dataset.html}{link}
    \item \textbf{MSMT-17} \href{https://www.pkuvmc.com/dataset.html}{link}
    \item \textbf{CUHK02} \href{https://www.ee.cuhk.edu.hk/~xgwang/CUHK_identification.html}{link}
    \item \textbf{CUHK03} \href{https://www.kaggle.com/datasets/priyanagda/cuhk03?resource=download}{link}
    \item \textbf{GRID} \href{https://personal.ie.cuhk.edu.hk/~ccloy/downloads_qmul_underground_reid.html}{link}
    \item \textbf{iLIDS} \href{https://xiatian-zhu.github.io/downloads_qmul_iLIDS-VID_ReID_dataset.html}{link}
\end{itemize}

\item Code availability

We have attached our code in the supplementary material. The code will be publicly open upon acceptance. 
\end{itemize}









\bibliography{sn-bibliography}

\end{document}